\documentclass[conference,a4paper]{IEEEtran}
\IEEEoverridecommandlockouts
\usepackage{cite}
\usepackage{amsmath,amssymb,amsfonts}
\usepackage{algorithmic}
\usepackage{graphicx}
\usepackage{textcomp}
\usepackage[table,xcdraw]{xcolor}

\def\BibTeX{{\rm B\kern-.05em{\sc i\kern-.025em b}\kern-.08em
    T\kern-.1667em\lower.7ex\hbox{E}\kern-.125emX}}
    
\graphicspath{{images/}}
\usepackage{tabularx}
\usepackage[hidelinks]{hyperref}
\usepackage{orcidlink}
\usepackage{booktabs}
\usepackage{multirow}
\usepackage{array}
\usepackage{makecell}
\usepackage{colortbl}

\usepackage{multicol}  % For multiple columns
\usepackage{adjustbox} % For adjusting table size if needed
\usepackage{lipsum}    % For generating dummy text
\usepackage[caption=false,font=footnotesize]{subfig}

\newcommand\copyrighttext{%
    \footnotesize \textcopyright\ 2025 IEEE. Personal use of this material is permitted. Permission from IEEE must be obtained for all other uses, in any current or future media, including reprinting/republishing this material for advertising or promotional purposes, creating new collective works, for resale or redistribution to servers or lists, or reuse of any copyrighted component of this work in other works. \\
    This is the author's accepted manuscript of the paper published in the \textit{2025 International Conference on New Trends in Computing Sciences (ICTCS)}. DOI: \href{https://doi.org/10.1109/ICTCS65341.2025.10989390}{10.1109/ICTCS65341.2025.10989390}.
}

\newcommand\copyrightnotice{%
\begin{tikzpicture}[remember picture,overlay]
\node[anchor=south,yshift=10pt] at (current page.south) {\fbox{\parbox{\dimexpr\textwidth-\fboxsep-\fboxrule\relax}{\copyrighttext}}};
\end{tikzpicture}%
}

\begin{document}

%%%%%%%%%%%%%%%%%%%%%%%%%%%%%%%%%%%%%%%%%%%%%%%%%%%%%%%%%%%%%%%%%%%%%%%%%%%%%%%%%%%%%%%
%%%%%%%%%%%%%%%%%%%%%%%%%%%%%%%%%%%%% Title %%%%%%%%%%%%%%%%%%%%%%%%%%%%%%%%%%%%%%%%%%%
%%%%%%%%%%%%%%%%%%%%%%%%%%%%%%%%%%%%%%%%%%%%%%%%%%%%%%%%%%%%%%%%%%%%%%%%%%%%%%%%%%%%%%% 
\title{A Multi-Step Comparative Framework for Anomaly Detection in IoT Data Streams}
%%%%%%%%%%%%%%%%%%%%%%%%%%%%%%%%%%%%%%%%%%%%%%%%%%%%%%%%%%%%%%%%%%%%%%%%%%%%%%%%%%%%%%%
%%%%%%%%%%%%%%%%%%%%%%%%%%%%%%%%%%%%%%%%%%%%%%%%%%%%%%%%%%%%%%%%%%%%%%%%%%%%%%%%%%%%%%% 

%%%%%%%%%%%%%%%%%%%%%%%%%%%%%%%%%%%%%%%%%%%%%%%%%%%%%%%%%%%%%%%%%%%%%%%%%%%%%%%%%%%%%%%
%%%%%%%%%%%%%%%%%%%%%%%%%%%%%%%%%%%%% Authors %%%%%%%%%%%%%%%%%%%%%%%%%%%%%%%%%%%%%%%%%
%%%%%%%%%%%%%%%%%%%%%%%%%%%%%%%%%%%%%%%%%%%%%%%%%%%%%%%%%%%%%%%%%%%%%%%%%%%%%%%%%%%%%%% 
\author{
\IEEEauthorblockN{Mohammed Al-Qudah} %\orcidlink{0009-0000-2554-7270}}
\IEEEauthorblockA
{
    \textit{Information Technology Department}\\
    \textit{Alwasl University}\\
    Dubai, UAE\\
    Email: mohammed.alqudah@alwasl.ac.ae\\
    ORCID: \orcidlink{0009-0000-2554-7270} 0009-0000-2554-7270
}
\and
\IEEEauthorblockN{Fadi AlMahamid} %\orcidlink{0000-0002-6907-7626}}
\IEEEauthorblockA
{
    \textit{Department of Electrical and Computer Engineering}\\
    \textit{Western University}\\
    London, Ontario, Canada\\
    Email: fadi.almahamid@uwo.ca\\
    ORCID: \orcidlink{0000-0002-6907-7626} 0000-0002-6907-7626
}
}
%%%%%%%%%%%%%%%%%%%%%%%%%%%%%%%%%%%%%%%%%%%%%%%%%%%%%%%%%%%%%%%%%%%%%%%%%%%%%%%%%%%%%%%

%%%%%%%%%%%%%%%%%%%%%%%%%%%%%%%%%% Make Titke %%%%%%%%%%%%%%%%%%%%%%%%%%%%%%%%%%%%%%%%%

\maketitle
\copyrightnotice
%%%%%%%%%%%%%%%%%%%%%%%%%%%%%%%%%%%%%%%%%%%%%%%%%%%%%%%%%%%%%%%%%%%%%%%%%%%
%%%%%%%%%%%%%%%%%%%%%%%%%%%%%%%%%%%% Abstract %%%%%%%%%%%%%%%%%%%%%%%%%%%%%
%%%%%%%%%%%%%%%%%%%%%%%%%%%%%%%%%%%%%%%%%%%%%%%%%%%%%%%%%%%%%%%%%%%%%%%%%%%
\begin{abstract}
The rapid expansion of Internet of Things (IoT) devices has introduced critical security challenges, underscoring the need for accurate anomaly detection. Although numerous studies have proposed machine learning (ML) methods for this purpose, limited research systematically examines how different preprocessing steps--normalization, transformation, and feature selection--interact with distinct model architectures. To address this gap, this paper presents a multi-step evaluation framework assessing the combined impact of preprocessing choices on three ML algorithms: RNN-LSTM, autoencoder neural networks (ANN), and Gradient Boosting (GBoosting). Experiments on the IoTID20 dataset shows that GBoosting consistently delivers superior accuracy across preprocessing configurations, while RNN-LSTM shows notable gains with z-score normalization and autoencoders excel in recall, making them well-suited for unsupervised scenarios. By offering a structured analysis of preprocessing decisions and their interplay with various ML techniques, the proposed framework provides actionable guidance to enhance anomaly detection performance in IoT environments.
\end{abstract}
%%%%%%%%%%%%%%%%%%%%%%%%%%%%%%%%%%%%%%%%%%%%%%%%%%%%%%%%%%%%%%%%%%%%%%%%%%%%%%%%%%%%%%%
%%%%%%%%%%%%%%%%%%%%%%%%%%%%%%%%%%%%%%%%%%%%%%%%%%%%%%%%%%%%%%%%%%%%%%%%%%%%%%%%%%%%%%%

%%%%%%%%%%%%%%%%%%%%%%%%%%%%%%%%%%%%%%%%%%%%%%%%%%%%%%%%%%%%%%%%%%%%%%%%%%%%%%%%%%%%%%%
%%%%%%%%%%%%%%%%%%%%%%%%%%%%%%%%%%%% Keywords %%%%%%%%%%%%%%%%%%%%%%%%%%%%%%%%%%%%%%%%%
%%%%%%%%%%%%%%%%%%%%%%%%%%%%%%%%%%%%%%%%%%%%%%%%%%%%%%%%%%%%%%%%%%%%%%%%%%%%%%%%%%%%%%%
%%%%%%%%%%%%%%%%%%%%%%%%%%%%%%%%%%%%%%%%%%%%%%%%%%%%%%%%%%%%%%%%%%%%%%%%%%%%%%%%%%%%%%%
%%%%%%%%%%%%%%%%%%%%%%%%%%%%%%%%%%%%%%%%%%%%%%%%%%%%%%%%%%%%%%%%%%%%%%%%%%%%%%%%%%%%%%%

%%%%%%%%%%%%%%%%%%%%%%%%%%%%%%%%%%%%%%%%%%%%%%%%%%%%%%%%%%%%%%%%%%%%%%%%%%%%%%%%%%%%%%%
%%%%%%%%%%%%%%%%%%%%%%%%%%%%%%%%%%%%%%%%%%%%%%%%%%%%%%%%%%%%%%%%%%%%%%%%%%%%%%%%%%%%%%%
%%%%%%%%%%%%%%%%%%%%%%%%%%%%%%%%%%%%%%%%%%%%%%%%%%%%%%%%%%%%%%%%%%%%%%%%%%%%%%%%%%%%%%%
%%%%%%%%%%%%%%%%%%%%%%%%%%%%%%%%%%%%%%%%%%%%%%%%%%%%%%%%%%%%%%%%%%%%%%%%%%%%%%%%%%%%%%%
\section{Introduction} \label{sec:introduction}
%%%%%%%%%%%%%%%%%%%%%%%%%%%%%%%%%%%%%%%%%%%%%%%%%%%%%%%%%%%%%%%%%%%%%%%%%%%%%%%%%%%%%%%
%%%%%%%%%%%%%%%%%%%%%%%%%%%%%%%%%%%%%%%%%%%%%%%%%%%%%%%%%%%%%%%%%%%%%%%%%%%%%%%%%%%%%%%
The exponential growth of the Internet of Things (IoT) has fueled innovations across diverse sectors, including smart homes, healthcare, and manufacturing. However, the widespread adoption of IoT devices also introduces significant security risks. A notable example is the 2016 Mirai botnet attack~\cite{antonakakis2017understanding}, in which compromised IoT devices orchestrated a large-scale distributed denial-of-service (DDoS) assault, disrupting major internet service providers. This incident underscores the vulnerability of IoT networks, especially when devices lack robust security measures.

To mitigate such threats, it is crucial to develop accurate anomaly detection mechanisms for IoT data streams. Recent advances in machine learning (ML) have enabled powerful solutions capable of identifying subtle deviations in network traffic that may indicate malicious activities. Yet, the effectiveness of these approaches heavily depends on data preprocessing decisions--such as normalization, feature selection, and transformations--as well as on the choice of ML algorithm.

Although prior research has explored various anomaly detection methods--ranging from recurrent neural networks (RNNs) to autoencoders and tree-based ensembles—there is a notable shortage of systematic evaluations comparing these techniques under uniform experimental conditions. Most studies focus on either a single algorithm or minimal preprocessing changes, leaving open questions about how different pipeline decisions interact to influence overall detection performance in diverse IoT attack scenarios.

This paper proposes a comprehensive, multi-step evaluation framework that simultaneously addresses:
\begin{itemize}
    \item \textbf{Preprocessing Techniques:} The framework investigates the impact of two normalization approaches (z-score and MinMax) and the Yeo--Johnson transformation on anomaly detection.
    \item \textbf{Feature Selection Methods:} The framework integrates Chi-square (Chi2) and Recursive Feature Elimination with Cross-Validation (RFECV) to assess their impact on detection accuracy and complexity.
    \item \textbf{Multiple ML Models:} The framework benchmarks Recurrent Neural Networks with LSTM (RNN-LSTM), autoencoder neural networks (Autoencoders), and Gradient Boosting (GBoosting) under a single, consistent experimental setting.
\end{itemize}

A key aspect of this study is the \textbf{IoTID20 dataset}~\cite{ullah2020scheme}, chosen for its realistic capture of both benign and malicious IoT network traffic. IoTID20 encompasses various attack vectors and normal behaviors, making it a suitable testbed for evaluating the interplay between different data preprocessing steps and detection models. By performing all experiments under uniform conditions on this dataset, the proposed framework uniquely shows how particular pipeline decisions can amplify or hinder each algorithm’s effectiveness.

Through extensive experiments, this work identifies which model--preprocessing configurations excel at detecting anomalies and provides practical guidelines for improving IoT security. In contrast to prior efforts limited to one or two methods, our multi-step, single-dataset approach offers deeper insights into optimal anomaly detection strategies when normalizing or transforming data and selecting salient features. 

The remainder of the paper is organized as follows: Section~\ref{sec:background} clarifies the core ML techniques used in this study. Section~\ref{sec:related-work} examines the state-of-the-art approaches in IoT anomaly detection. Section~\ref{sec:methodology} details our proposed evaluation framework, including data preparation and feature selection. Section~\ref{sec:evaluation} presents and discusses the experimental results. Finally, Section~\ref{sec:conclusion} concludes the paper and outlines potential directions for future research.

%%%%%%%%%%%%%%%%%%%%%%%%%%%%%%%%%%%%%%%%%%%%%%%%%%%%%%%%%%%%%%%%%%%%%%%%%%%%%%%%%%%%%%%
%%%%%%%%%%%%%%%%%%%%%%%%%%%%%%%%%%%%%%%%%%%%%%%%%%%%%%%%%%%%%%%%%%%%%%%%%%%%%%%%%%%%%%% 
\section{Background} \label{sec:background}
%%%%%%%%%%%%%%%%%%%%%%%%%%%%%%%%%%%%%%%%%%%%%%%%%%%%%%%%%%%%%%%%%%%%%%%%%%%%%%%%%%%%%%%
%%%%%%%%%%%%%%%%%%%%%%%%%%%%%%%%%%%%%%%%%%%%%%%%%%%%%%%%%%%%%%%%%%%%%%%%%%%%%%%%%%%%%%% 
This section provides an overview of the key concepts and techniques used in the evaluation framework. It first introduces three primary machine learning approaches for IoT anomaly detection: \textit{Recurrent Neural Networks with LSTM}, \textit{Autoencoder Neural Networks}, and \textit{Gradient Boosting}. It then outlines essential data preprocessing steps, including normalization, transformation, and feature selection.
%=====================================================================================%
\subsection{Machine Learning Approaches for IoT Anomaly Detection}
%=====================================================================================%

%-------------------------------------------------------------------------------------%
\subsubsection{RNN-LSTM}
%-------------------------------------------------------------------------------------%
Recurrent Neural Networks (RNNs) are well-suited for sequence modeling, as they allow information to persist across time steps. The Long Short-Term Memory (LSTM) variant is designed to overcome the vanishing gradient problem typically encountered with standard RNNs. LSTMs can capture both short and long-term dependencies in time-series data. In the context of IoT, LSTM layers help detect anomalies in network traffic over time by learning temporal patterns indicative of normal versus malicious behavior.

%-------------------------------------------------------------------------------------%
\subsubsection{Autoencoders}
%-------------------------------------------------------------------------------------%
Autoencoders are neural networks trained to reconstruct their own inputs through an encoder–decoder architecture. During training, the network learns a compressed latent representation of the data. In anomaly detection tasks, autoencoders can model the normal data distribution. If a new sample reconstructs poorly, it is flagged as anomalous. This approach can be especially beneficial in unsupervised or semi-supervised settings where labeled attack data is scarce.

%-------------------------------------------------------------------------------------%
\subsubsection{Gradient Boosting}
%-------------------------------------------------------------------------------------%
Gradient Boosting (GBoosting) constructs an ensemble of weak learners (typically decision trees) in a stage-wise manner. Each new learner attempts to correct the residuals of the previous ensemble, resulting in a powerful final model. In IoT traffic analysis, boosting algorithms often excel at handling diverse feature types and can achieve high accuracy with systematic hyperparameter tuning. Their ability to iteratively reduce errors makes them robust to complex data distributions frequently observed in IoT systems.

%=====================================================================================%
\subsection{Data Preprocessing Techniques}
%=====================================================================================%
%-------------------------------------------------------------------------------------%
\subsubsection{Normalization}
%-------------------------------------------------------------------------------------%
Normalization reduces biases caused by large numeric ranges in feature values, improving training stability. Two common methods are:
\begin{itemize}
    \item \textbf{z-score (Standardization):} Scales the data to zero mean and unit variance. It is especially useful for features assumed to follow an approximately normal distribution.
    \item \textbf{MinMax Scaling:} Scales the data to a fixed range, often [0, 1]. This is helpful when all features share a known minimum and maximum.
\end{itemize}

%-------------------------------------------------------------------------------------%
\subsubsection{Transformation}
%-------------------------------------------------------------------------------------%
Data transformation addresses skewed distributions or outlier effects \cite{almujally2024biosensor}. The Yeo--Johnson power transformation can handle both positive and negative values without shifting the data. Transformations may improve accuracy by making feature distributions close to Gaussian distribution. However, the benefit varies depending on the underlying data and the chosen learning algorithm.

%-------------------------------------------------------------------------------------%
\subsubsection{Feature Selection}
%-------------------------------------------------------------------------------------%
Feature selection aims to identify the most relevant subset of features, reducing training time and potentially enhancing model performance. Two widely used methods include:
\begin{itemize}
    \item \textbf{Recursive Feature Elimination with Cross-Validation (RFECV):} Iteratively ranks features by importance and eliminates those contributing least to the model, retraining at each step to find an optimal subset.\cite{awad2023recursive}
    \item \textbf{Chi-square (Chi2) Test:} Evaluates each feature’s statistical relationship with the target class. Higher Chi2 values suggest a stronger correlation.\cite{alghazzawi2021efficient} Since Chi2 requires non-negative data, it is often combined with a non-negative normalization approach.
\end{itemize}

Overall, the combination of normalization, transformation, and feature selection can significantly influence the effectiveness of anomaly detection models. The subsequent sections elaborate on how these preprocessing steps integrate with the ML techniques described above.

%%%%%%%%%%%%%%%%%%%%%%%%%%%%%%%%%%%%%%%%%%%%%%%%%%%%%%%%%%%%%%%%%%%%%%%%%%%%%%%%%%%%%%%
%%%%%%%%%%%%%%%%%%%%%%%%%%%%%%%%%%%%%%%%%%%%%%%%%%%%%%%%%%%%%%%%%%%%%%%%%%%%%%%%%%%%%%% 
\section{Related Work} \label{sec:related-work}
%%%%%%%%%%%%%%%%%%%%%%%%%%%%%%%%%%%%%%%%%%%%%%%%%%%%%%%%%%%%%%%%%%%%%%%%%%%%%%%%%%%%%%%
%%%%%%%%%%%%%%%%%%%%%%%%%%%%%%%%%%%%%%%%%%%%%%%%%%%%%%%%%%%%%%%%%%%%%%%%%%%%%%%%%%%%%%% 
Researchers have proposed a range of ML-driven solutions for detecting anomalies in IoT data, with methods spanning deep learning architectures to tree-based models. This section summarizes existing approaches and highlights complementary studies on IoT data streams, comparative evaluations, and preprocessing techniques to contextualize our comprehensive framework.

\noindent
\textbf{Autoencoders}\\
Lopez-Martin et al.~\cite{lopez2017conditional} introduced a conditional variational autoencoder (VAE) for IoT intrusion detection, integrating labels into the decoder layer to better reconstruct missing information. Similarly, Abusitta et al.~\cite{abusitta2023deep} implemented a denoising autoencoder as a feature extraction step, subsequently feeding these features into an SVM classifier to attain 94.6\% accuracy in detecting malicious traffic. Both studies demonstrate how autoencoders model normal behavior for outlier detection and underscore their potential for anomaly identification under diverse IoT conditions.

\noindent
\textbf{RNN-LSTM}\\
Park et al.~\cite{park2020rnn} explored an RNN-LSTM model that flags anomalies when observed time-series data deviates from the model's predictions. Hwang et al.~\cite{9024425} combined a convolutional neural network (CNN) with an LSTM-based recurrent autoencoder, leveraging a two-stage sliding window to capture spatial and temporal features. Although these approaches often achieve high accuracy, they do not systematically analyze how different data preprocessing choices affect performance across multiple ML algorithms.

\noindent
\textbf{GBoosting Models}\\
Douiba et al.~\cite{douiba2023anomaly} employed CatBoost, a specialized gradient boosting algorithm optimized for categorical data, to achieve over 99.9\% accuracy in detecting IoT anomalies. Bui et al.~\cite{bui2021gradient} further illustrated Gboosting's adaptability by applying it successfully to land-cover classification, indicating the robustness of tree-based methods in varied domains. Despite these promising results, limited work has compared Gboosting to alternative ML models under uniformly varied preprocessing settings in IoT contexts.

\noindent
\textbf{IoT Frameworks and Comparative Analyses}\\
Yang et al.~\cite{yang2022multi,yang2021pwpae} presented multi-stage frameworks emphasizing real-time processing and adaptive learning to handle evolving IoT data streams (concept drift). While these approaches highlight the value of flexible, multi-step solutions, they do not provide a uniform evaluation of diverse preprocessing techniques within a single experimental setting. Meanwhile, Inuwa and Das~\cite{inuwa2024comparative} offered a comparative assessment of ML models for cyberattack detection in IoT networks, underscoring the algorithmic breadth available. Nassif et al.~\cite{nassif2021machine} provided a systematic review of anomaly detection methods, detailing various ML techniques and their strengths. However, neither study focuses on how the combination of normalization, transformation, and feature selection might affect model performance when tested under consistent conditions.

\noindent
\textbf{Motivation and Gap}\\
While these studies underscore the efficacy of autoencoders, RNN-LSTM, and Gboosting for IoT anomaly detection, the collective literature seldom compares how different preprocessing decisions--normalization, transformations, feature selection--interact with multiple ML models on a single IoT dataset. This study addresses that gap by systematically examining how varied preprocessing pipelines affect model performance (autoencoders, RNN-LSTM, gradient boosting), aiming to provide deeper insights into optimal anomaly detection strategies for IoT systems.

%%%%%%%%%%%%%%%%%%%%%%%%%%%%%%%%%%%%%%%%%%%%%%%%%%%%%%%%%%%%%%%%%%%%%%%%%%%%%%%%%%%%%%%
%%%%%%%%%%%%%%%%%%%%%%%%%%%%%%%%%%%%%%%%%%%%%%%%%%%%%%%%%%%%%%%%%%%%%%%%%%%%%%%%%%%%%%%
\section{Methodology} \label{sec:methodology}
\begin{figure}[!t]
    \centering
    \includegraphics[width=.85\linewidth]{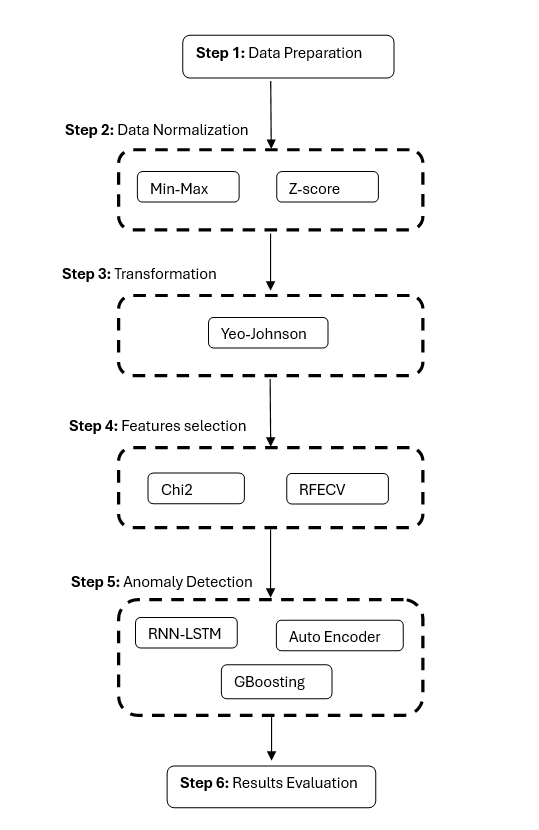}
    \caption{Proposed multi-step anomaly detection methodology in IoT environments}
    \label{fig:framework-methodology}
\end{figure}

%%%%%%%%%%%%%%%%%%%%%%%%%%%%%%%%%%%%%%%%%%%%%%%%%%%%%%%%%%%%%%%%%%%%%%%%%%%%%%%%%%%%%%%
%%%%%%%%%%%%%%%%%%%%%%%%%%%%%%%%%%%%%%%%%%%%%%%%%%%%%%%%%%%%%%%%%%%%%%%%%%%%%%%%%%%%%%%
Figure~\ref{fig:framework-methodology} illustrates the multi-step process proposed for detecting anomalies in IoT data. The methodology consists of six primary steps: Data Preparation, Data Normalization, Transformation, Features Selection, Anomaly Detection, and Results Evaluation. Each step is discussed in the following subsections.

%=====================================================================================%
\subsection{Data Preparation}
%=====================================================================================%
The IoTID20 dataset~\cite{ullah2020scheme} is employed for anomaly detection, containing both normal and malicious IoT network traffic. It includes diverse flow-based features such as packet sizes, protocol information, and timestamps. The dataset is first checked for missing values, invalid entries, or duplicates, which are removed to ensure data quality. 

Each record in the dataset represents a traffic flow labeled as either normal or malicious. After cleaning, the final dataset is split into 80\% for training and 20\% for testing. This split ensures that the trained models are evaluated on unseen data, providing an estimate of their generalization performance.

%=====================================================================================%
\subsection{Data Normalization}
%=====================================================================================%
Normalization mitigates biases arising from large or varying feature scales. In this work, two common techniques are applied:

\begin{itemize}
    \item \textbf{Min-Max Scaling:} 
    Rescales each feature to the range [0, 1]:
    \begin{equation}
        x_{\mathrm{scaled}} = \frac{x - \min(x)}{\max(x) - \min(x)}
        \label{eq:minmax}
    \end{equation}
    where \(\min(x)\) and \(\max(x)\) are the minimum and maximum values of a feature in the training set.

    \item \textbf{z-score Normalization:} 
    Standardizes each feature to zero mean and unit variance:
    \begin{equation}
        z = \frac{x - \overline{x}}{\sigma}
        \label{eq:z-score}
    \end{equation}
    where \(\overline{x}\) is the mean and \(\sigma\) is the standard deviation of the feature in the training set.
\end{itemize}

All normalization parameters (mean, standard deviation, min, max) are computed exclusively on the training set to prevent data leakage. These parameters are then applied to scale the test set, emulating real-world conditions where future data distributions remain unknown at training time. 

%=====================================================================================%
\subsection{Transformation}
%=====================================================================================%
When feature distributions exhibit skewness or include negative values, transformations can enhance model robustness\cite{weisberg2001yeo}. The Yeo–Johnson power transformation is optionally employed in this methodology. It handles both positive and negative feature values without requiring a data shift:

\begin{equation}
y(\lambda) =
\begin{cases}
    \frac{(x + 1)^\lambda - 1}{\lambda}, & \text{if } \lambda \neq 0,\; x \geq 0,\\
    \log(x + 1), & \text{if } \lambda = 0,\; x \geq 0,\\
    -\frac{(-x + 1)^{2 - \lambda} - 1}{2 - \lambda}, & \text{if } \lambda \neq 2,\; x < 0,\\
    -\log(-x + 1), & \text{if } \lambda = 2,\; x < 0.
\end{cases}
\label{eq:yeo-johnson}
\end{equation}

A suitable \(\lambda\) parameter is determined based on the training set, often via maximum likelihood estimation or cross-validation. The test set is transformed using the same \(\lambda\). Only configurations requiring this transformation apply it during experimentation.

%=====================================================================================%
\subsection{Features Selection}
%=====================================================================================%
Selecting an optimal subset of features can improve classification performance and reduce computational complexity. Two feature selection methods are considered:

\begin{itemize}
    \item \textbf{Recursive Feature Elimination with Cross-Validation (RFECV):}  
    This process iteratively ranks features based on their importance to a chosen model and removes those deemed least critical. Cross-validation is employed to determine the best number of features to retain.

    \item \textbf{Chi-square (Chi2) Test:}  
    This statistical test assesses whether each feature is independent of the anomaly label. Features with higher Chi2 values have stronger relevance. Because Chi2 requires non-negative data, it is typically combined with MinMax scaling.
\end{itemize}

%=====================================================================================%
\subsection{Anomaly Detection}
%=====================================================================================%
After preprocessing, three machine-learning models are deployed to detect anomalies:

\begin{itemize}
    \item \textbf{RNN-LSTM:} 
    A time-series neural network capable of modeling temporal dependencies in IoT traffic. The architecture is configured to capture patterns that differentiate benign from malicious behaviors.
    \item \textbf{Autoencoder:} 
    An unsupervised approach trained to reconstruct normal traffic with minimal error. A high reconstruction error often indicates an anomalous input.
    \item \textbf{Gradient Boosting (GBoosting):} 
    A tree-ensemble algorithm that iteratively refines weak learners to minimize classification errors. It is effective for heterogeneous feature types and can achieve high accuracy with proper tuning.
\end{itemize}

Each model is trained using one or more combinations of the above-mentioned preprocessing steps (i.e., normalization, transformation, and feature selection). Hyperparameters (e.g., learning rate, number of hidden units, number of estimators) are fine-tuned via cross-validation on the training set.

%=====================================================================================%
\subsection{Results Evaluation}
%=====================================================================================%
The final step measures the performance of each model-preprocessing combination using four standard classification metrics:

\begin{itemize}
    \item \textbf{Accuracy:} Proportion of correctly classified samples among all test samples.
    \item \textbf{Precision:} The Ratio of true positives to all predicted positives reflects how many flagged anomalies are genuinely malicious.
    \item \textbf{Recall:} The Ratio of true positives to all actual anomalies indicates how many malicious activities are correctly detected.
    \item \textbf{F1-score:} The harmonic mean of precision and recall offers a balanced metric for imbalanced data.
\end{itemize}

These metrics are reported on the 20\% test set that was held out during training. By systematically varying the normalization (z-score or Min-Max), transformation (Yeo–Johnson or none), and feature selection (RFECV, Chi2, or none), this methodology identifies which configurations maximize anomaly detection accuracy and minimize false alarms in IoT networks. Subsequent sections present and discuss the experimental outcomes.

%%%%%%%%%%%%%%%%%%%%%%%%%%%%%%%%%%%%%%%%%%%%%%%%%%%%%%%%%%%%%%%%%%%%%%%%%%%%%%%%%%%%%%%
%%%%%%%%%%%%%%%%%%%%%%%%%%%%%%%%%%%%%%%%%%%%%%%%%%%%%%%%%%%%%%%%%%%%%%%%%%%%%%%%%%%%%%%
\section{Evaluation}\label{sec:evaluation}
%%%%%%%%%%%%%%%%%%%%%%%%%%%%%%%%%%%%%%%%%%%%%%%%%%%%%%%%%%%%%%%%%%%%%%%%%%%%%%%%%%%%%%%
%%%%%%%%%%%%%%%%%%%%%%%%%%%%%%%%%%%%%%%%%%%%%%%%%%%%%%%%%%%%%%%%%%%%%%%%%%%%%%%%%%%%%%%
Multiple experiments were conducted using various combinations of normalization, transformation, and feature selection methods on the IoTID20 dataset to evaluate their impact on anomaly detection performance systematically. Specifically, three machine learning models (GBoosting, RNN-LSTM, and autoencoder) were assessed across these preprocessing configurations. Four performance metrics (accuracy, precision, recall, and F1-score) were employed to comprehensively analyze each model's effectiveness and highlight strengths and limitations.

Table~\ref{tab:performance-comp} summarizes the performance results for all considered models and preprocessing combinations, providing a comparative overview.

\begin{table*}[t!]
\caption{Performance Comparison of Methods Grouped by Anomaly Detection Technique}
\label{tab:performance-comp}
\centering
\begin{tabular}{llllllllll}
\toprule
\rowcolor[HTML]{EFEFEF}
\textbf{Model} & \textbf{ID} & \textbf{Normalization} & \textbf{Transformation} & \textbf{Features Selection} &  \textbf{Features \#} & \textbf{Accuracy} & \textbf{Precision} & \textbf{Recall} & \textbf{F1-Score} \tabularnewline

\midrule
\multirow{12}{*}{RNN-LSTM} 
  & 1  & min-max scaling & Yeo-Johnson & Chi2  & 20 & 99.28\% & 99.41\% & 99.83\% & 99.62\% \tabularnewline
  & 2  & min-max scaling & Yeo-Johnson & RFECV & 1  & 94.32\% & 94.32\% & 100.00\% & 97.08\% \tabularnewline
  & 3  & min-max scaling & Yeo-Johnson & N/A   & 31  & 99.04\% & 99.08\% & 99.92\% & 99.49\% \tabularnewline
  & 4  & min-max scaling & N/A         & Chi2  & 20 & 95.36\% & 96.98\% & 98.14\% & 97.56\% \tabularnewline
  & 5  & min-max scaling & N/A         & RFECV & 16 & 98.00\% & 98.82\% & 99.07\% & 98.94\% \tabularnewline
  & 6  & min-max scaling & N/A         & N/A   & 31 & 97.84\% & 98.57\% & 99.15\% & 98.86\% \tabularnewline
  & 7  & z-score      & Yeo-Johnson & RFECV & 23 & 98.32\% & 98.99\% & 99.24\% & 99.11\% \tabularnewline
  & 8  & z-score      & Yeo-Johnson & N/A   & 31 & 99.20\% & 99.49\% & 99.66\% & 99.58\% \tabularnewline
  & 9  & z-score      & N/A         & RFECV & 25 & 99.28\% & 99.49\% & 99.75\% & 99.62\% \tabularnewline
  & 10 & z-score      & N/A         & N/A   & 31 & \cellcolor[HTML]{40E0D0}  99.52\% & \cellcolor[HTML]{40E0D0} 99.58\% & 99.92\% & \cellcolor[HTML]{40E0D0} 99.75\% \tabularnewline
  & 11 & N/A          & Yeo-Johnson & N/A   & 31  & 94.32\% & 94.32\% &  \cellcolor[HTML]{40E0D0} 100.00\% & 97.08\% \tabularnewline
  & 12 & N/A          & N/A         & N/A   & 31 & 94.48\% & 95.35\% & 98.98\% & 97.13\% \tabularnewline
\midrule
\multirow{12}{*}{Autoencoder} 
  & 13 & min-max scaling & Yeo-Johnson & Chi2  & 20 & 93.29\% & 94.26\% & 98.90\% & 96.53\% \tabularnewline
  & 14 & min-max scaling & Yeo-Johnson & RFECV & 1 & 51.88\% & 93.52\% & 52.63\% & 67.35\% \tabularnewline
  & 15 & min-max scaling & Yeo-Johnson & N/A   & 31 & 93.29\% & 94.26\% & 98.90\% & 96.53\% \tabularnewline
  & 16 & min-max scaling & N/A         & Chi2  & 20 & 93.29\% & 94.26\% & 98.90\% & 96.53\% \tabularnewline
  & 17 & min-max scaling & N/A         & RFECV & 16 & 93.29\% & 94.26\% & 98.90\% & 96.53\% \tabularnewline
  & 18 & min-max scaling & N/A         & N/A   & 31 & 93.45\% & 94.35\% & 98.98\% & 96.61\% \tabularnewline
  & 19 & z-score      & Yeo-Johnson & RFECV & 23 & 93.45\% & 94.35\% & 98.98\% & 96.61\% \tabularnewline
  & 20 & z-score      & Yeo-Johnson & N/A   & 31 & 93.92\% & 94.59\% & 99.24\% & 96.86\% \tabularnewline
  & 21 & z-score      & N/A         & RFECV & 25 & 93.61\% & 94.43\% & 99.07\% & 96.69\% \tabularnewline
  & 22 & z-score      & N/A         & N/A   & 31 & \cellcolor[HTML]{FFA500} 94.08\% & \cellcolor[HTML]{FFA500} 94.67\% & \cellcolor[HTML]{FFA500} 99.32\% & \cellcolor[HTML]{FFA500} 96.94\% \tabularnewline
  & 23 & N/A          & Yeo-Johnson & N/A   & 31 & 93.29\% & 94.26\% & 98.90\% & 96.53\% \tabularnewline
  & 24 & N/A          & N/A         & N/A   & 31 & 93.29\% & 94.26\% & 98.90\% & 96.53\% \tabularnewline
\midrule
\multirow{10}{*}{GBoosting} 
  & 25 & min-max scaling & Yeo-Johnson & Chi2  & 20 & 99.84\% & 99.92\% & 99.92\% & 99.92\% \tabularnewline
  & 26 & min-max scaling & Yeo-Johnson & RFECV & 1  & 95.12\% & 95.90\% & 99.07\% & 97.46\% \tabularnewline
  & 27 & min-max scaling & Yeo-Johnson & N/A   & 31  & 99.84\% & 99.92\% & 99.92\% & 99.92\% \tabularnewline
  & 28 & min-max scaling & N/A         & Chi2  & 20 & 99.60\% & 99.66\% & 99.92\% & 99.79\% \tabularnewline
  & 29 & min-max scaling & N/A         & RFECV & 16 & 99.84\% & 99.92\% & 99.92\% & 99.92\% \tabularnewline
  & 30 & min-max scaling & N/A         & N/A   & 31  & 99.84\% & 99.92\% & 99.92\% & 99.92\% \tabularnewline
  & 31 & z-score      & Yeo-Johnson & N/A   & 31 & 99.84\% & 99.92\% & 99.92\% & 99.92\% \tabularnewline
  & 32 & z-score      & N/A         & N/A   & 31  & \cellcolor[HTML]{FF00FF} 99.84\% & \cellcolor[HTML]{FF00FF} 99.92\% & \cellcolor[HTML]{FF00FF} 99.92\% & \cellcolor[HTML]{FF00FF} 99.92\% \tabularnewline
  & 33 & N/A          & Yeo-Johnson & N/A   & 31  & 99.84\% & 99.92\% & 99.92\% & 99.92\% \tabularnewline
  & 34 & N/A          & N/A         & N/A   & 31  & 99.84\% & 99.92\% & 99.92\% & 99.92\% \tabularnewline
\bottomrule
\end{tabular}
\end{table*}

The subsequent subsections first describe the experimental setup, detailing the dataset partitioning, preprocessing strategies, and model architectures utilized in this research. Next, a detailed analysis presents the evaluation of how each preprocessing method impacts anomaly detection performance. Finally, the performance of individual machine learning models is analyzed, highlighting their respective strengths, sensitivities, and practical implications for IoT anomaly detection.
%=====================================================================================%
\subsection{Experimental Setup}
%=====================================================================================%
Experiments were consistently conducted using an 80/20 split to evaluate preprocessing impacts accurately for training and testing the IoTID20 dataset. The following model configurations were used:

\noindent
\textbf{RNN-LSTM Settings:}\\  
The RNN-LSTM architecture consisted of a single hidden layer containing 32 LSTM units with dropout regularization of 0.2, followed by a dense output layer for binary classification. The model was optimized using the Adam optimizer with a learning rate of 0.001 and trained for 30 epochs with a batch size of 32.

\noindent
\textbf{Autoencoder Settings:}\\
The autoencoder architecture comprised a symmetric design featuring an encoder layer (with ReLU activation) compressing the input to a reduced dimensional representation and a decoder layer (with ReLU activation) reconstructing the original input. Specifically, the encoder layer utilized 16 hidden units. The autoencoder was trained using the Adam optimizer with a learning rate of 0.001, binary cross-entropy as the loss function, and a batch size of 32 over 30 epochs.

\noindent
\textbf{GBoosting Settings:}\\  
The GBoosting model employed 100 estimators, a learning rate of 0.1, and a maximum depth of 3 to balance complexity and accuracy. The classification was performed using log-loss (deviance) optimization, allowing the model to effectively handle binary anomaly detection tasks.

The next subsection details the impact of each data preprocessing step on model performance.

%=====================================================================================%
\subsection{Impact Analysis of Data Preprocessing}
%=====================================================================================%
This section discusses how various data preprocessing techniques—specifically normalization, transformation, and feature selection—affected the anomaly detection performance of the selected machine learning algorithms. Understanding these impacts allows practitioners to make informed choices about preprocessing strategies that enhance the accuracy and efficiency of the model.

\begin{figure}[t!]
    \centering
    \subfloat[Before Normalization]{
        \includegraphics[width=.82\linewidth]{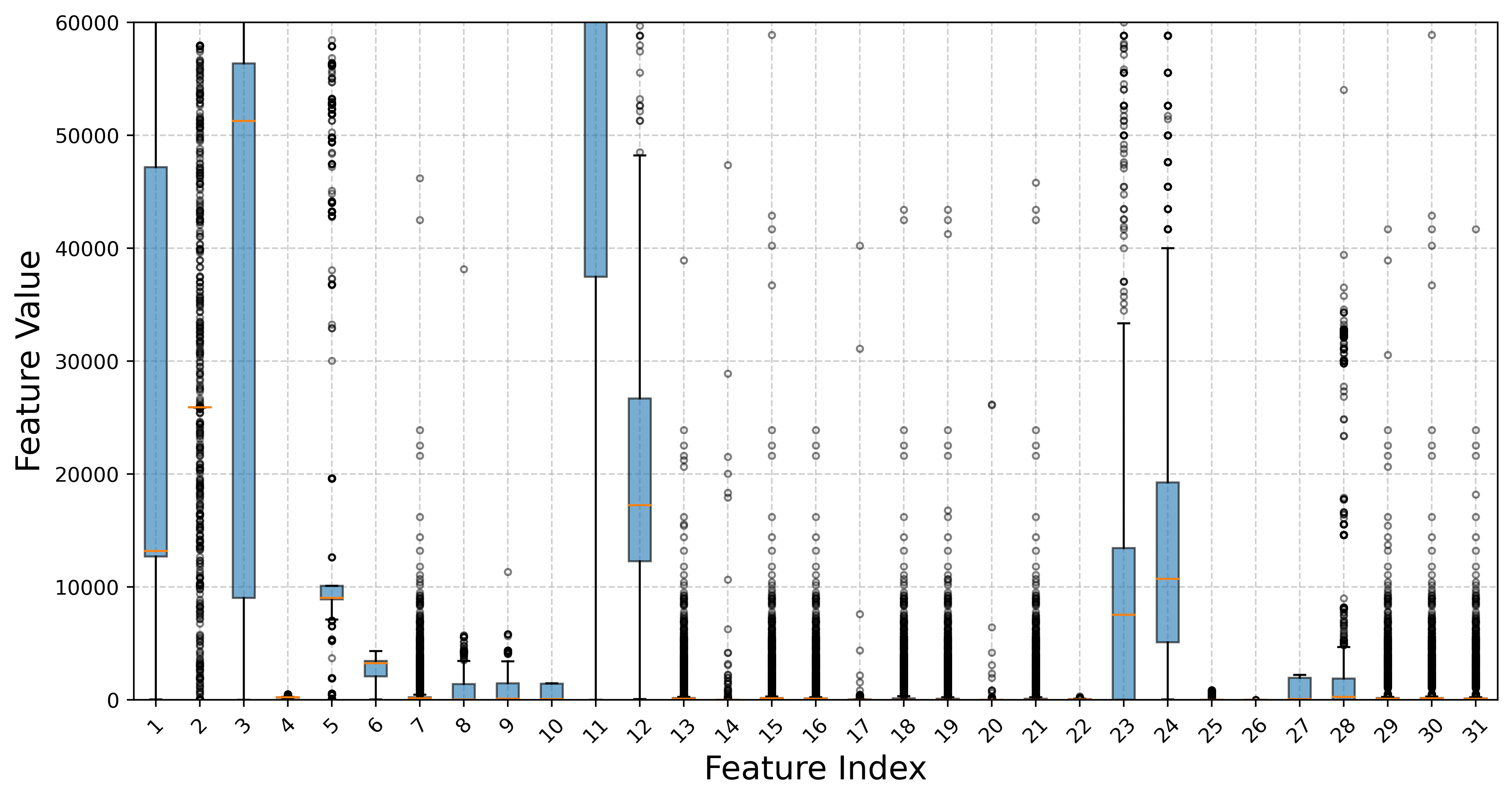}
        \label{fig:before-norm}
    }\\
    \subfloat[After min-max Scaling]{
        \includegraphics[width=.82\linewidth]{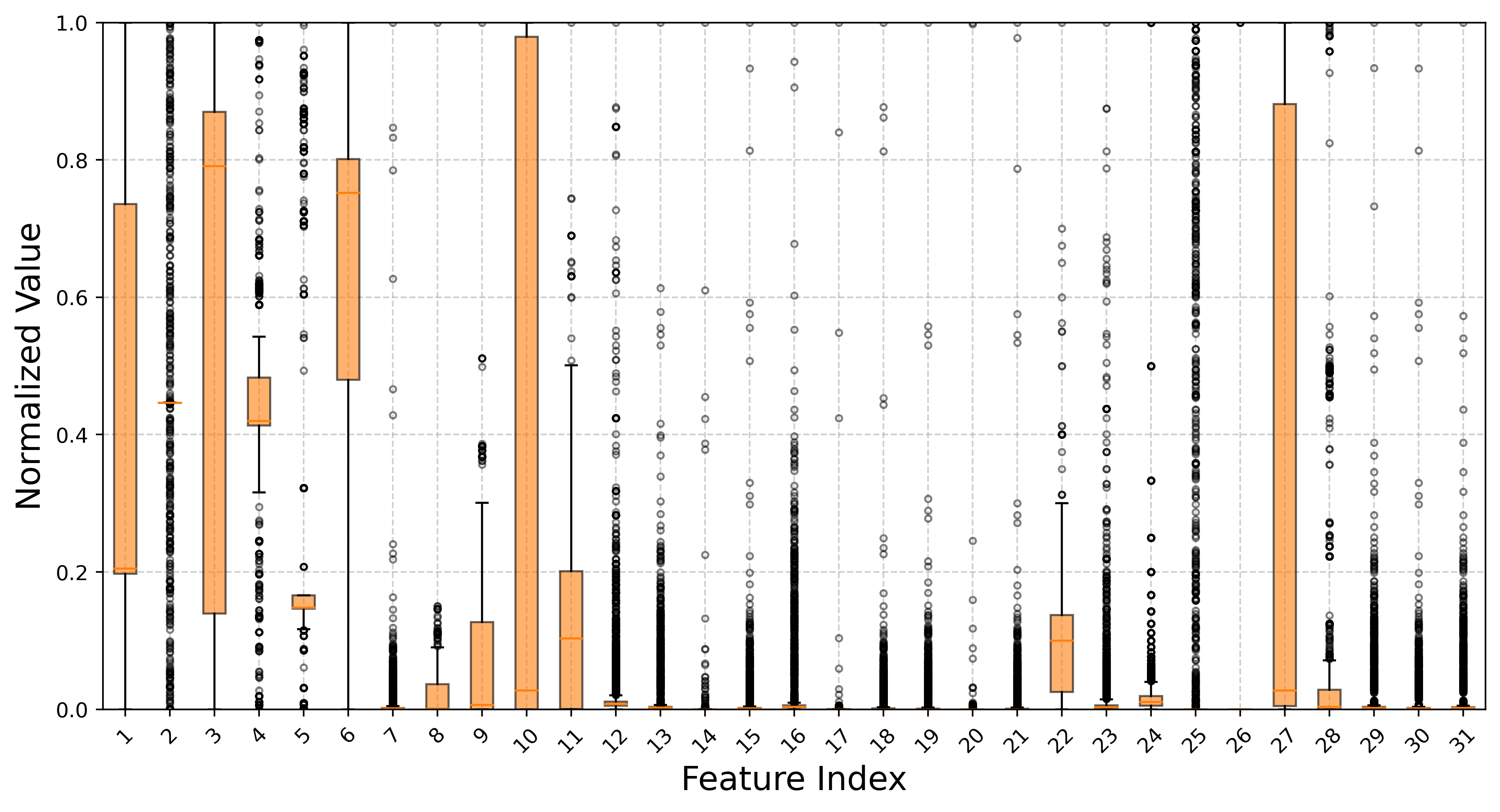}
        \label{fig:after-minmax}
    }\\
    \subfloat[After z-score Normalization]{
        \includegraphics[width=.82\linewidth]{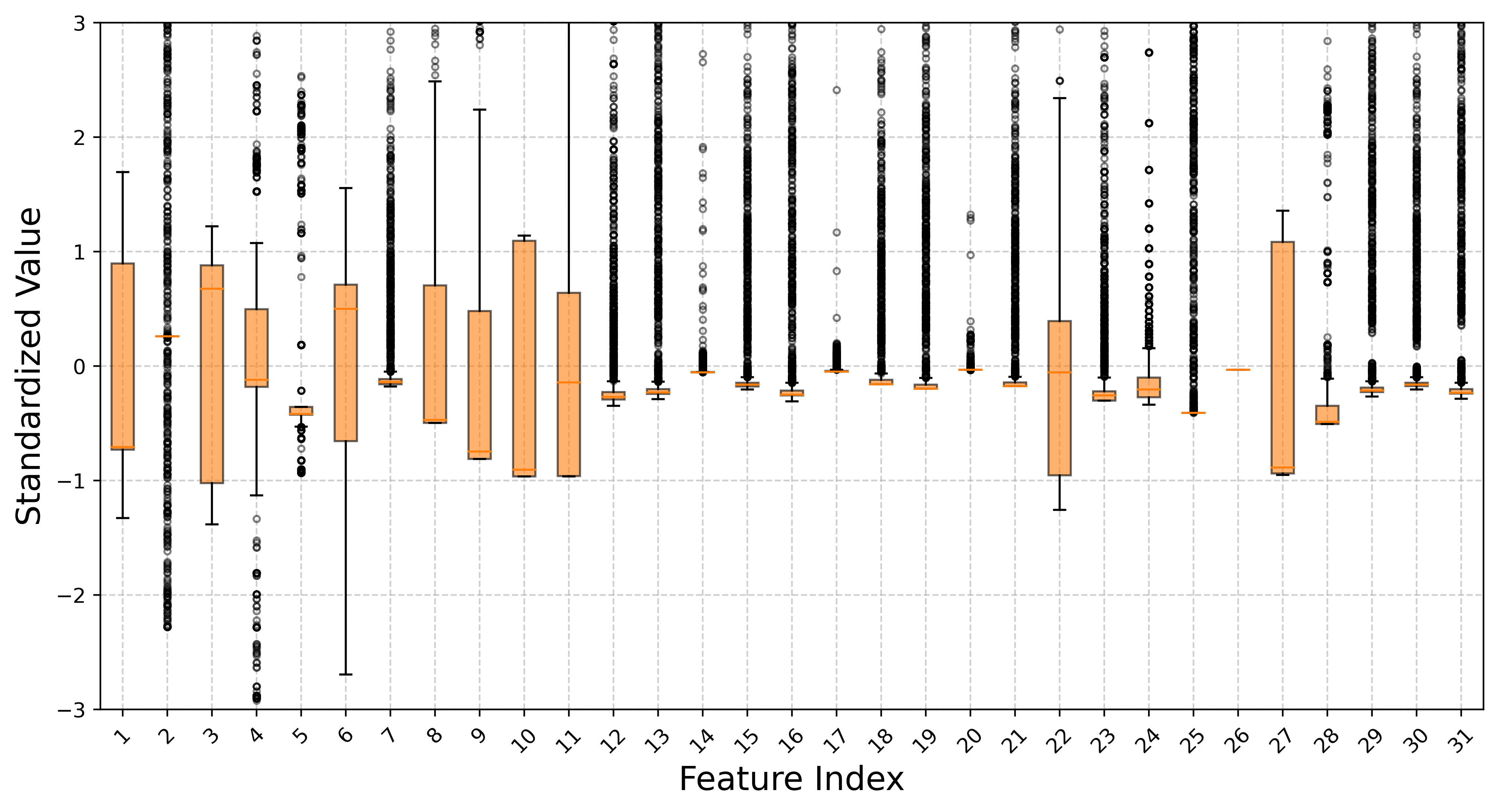}
        \label{fig:after-zscore}
    }
    \caption{Feature distributions before and after applying min-max scaling and z-score normalization.}
    \label{fig:normalization-impact}
\end{figure}

\noindent
\textbf{Normalization Impact}\\
Normalization showed pronounced effects across the evaluated models, underscoring the importance of choosing an appropriate scaling strategy. Initially, the dataset exhibited features with wide value ranges—some exceeding 50{,}000—as illustrated in Figure~\ref{fig:normalization-impact}-a.

Figure~\ref{fig:normalization-impact}-a highlights how raw data span large numeric ranges, potentially biasing algorithms sensitive to feature magnitudes. Applying min-max scaling rescales each feature to the interval $[0,1]$, as shown in Figure~\ref{fig:normalization-impact}-b. Although this transformation preserves relative differences among features, it may not sufficiently mitigate the influence of outliers. In several models (e.g., models 1 and 3 for RNN-LSTM), min-max scaling yielded moderate gains, but certain outlier-heavy features still impacted performance.

In contrast, z-score normalization as depicted in Figure~\ref{fig:normalization-impact}-c centers each feature around zero mean and unit variance, effectively reducing the effects of extreme values. Models such as RNN-LSTM (model~10) and Autoencoder (model~22) particularly benefited from this standardized data range, exhibiting higher accuracy or recall. By minimizing skewness and making features more comparable in scale, z-score normalization emerged as a consistently stronger option for anomaly detection in the selected dataset.

Overall, while both min-max scaling and z-score normalization improved model accuracy over raw, unscaled data, **z-score** showed stronger and more consistent improvements—especially for models that rely on well-behaved feature distributions (e.g., neural networks). Tree-based approaches, such as GBoosting, appeared more robust to either scaling strategy, displaying minimal performance deviations regardless of the chosen normalization method.

\noindent
\textbf{Transformation Impact}\\
The Yeo-Johnson transformation had minimal or negligible effects across the evaluated models. Its limited impact suggests that, for the IoTID20 dataset, additional transformations provided minimal value, implying simpler preprocessing pipelines may be sufficient.

\begin{figure}[t!]
    \centering
	\subfloat[Before Feature Selection]{
		\includegraphics[width=.9\linewidth]{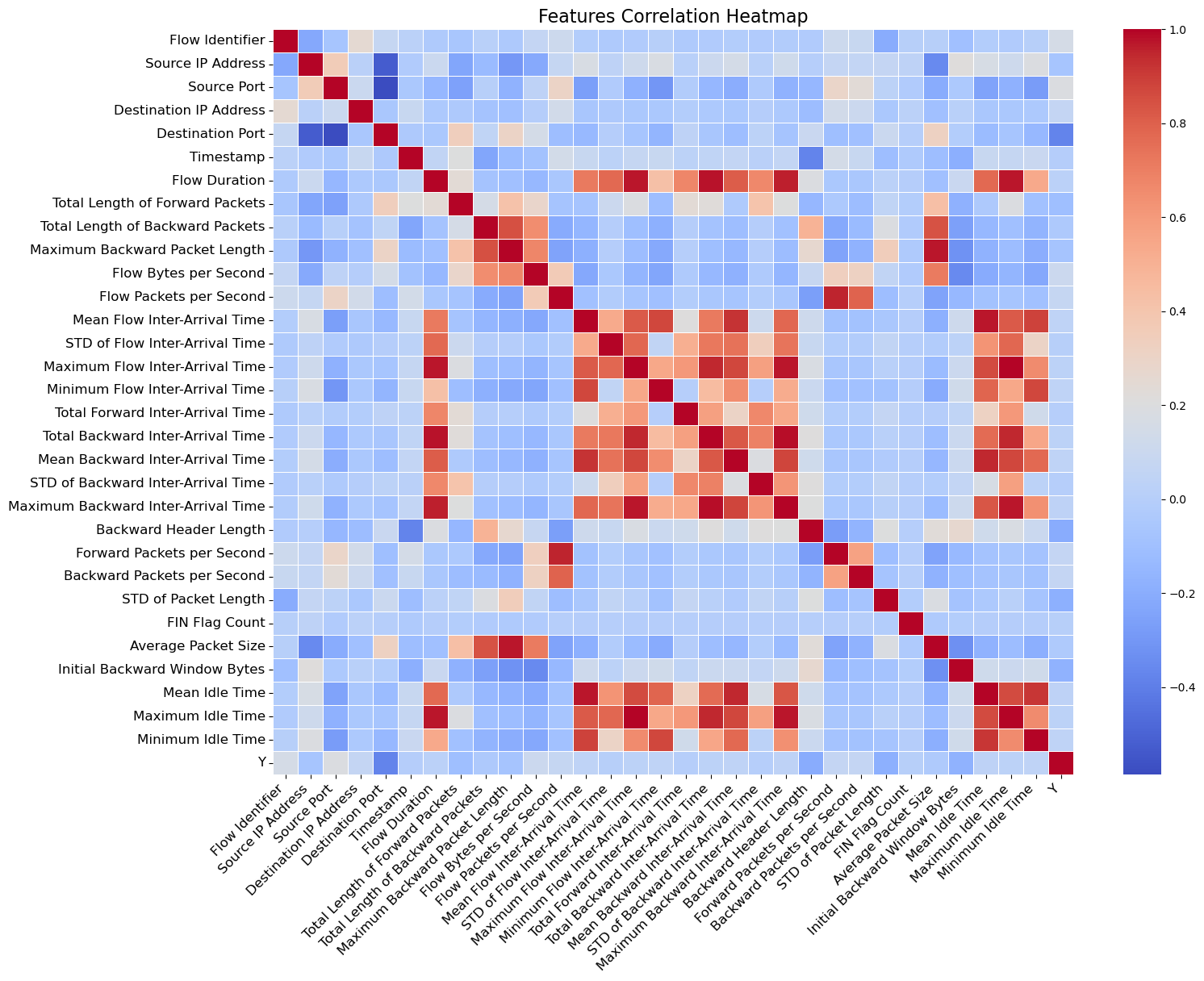}
		\label{fig:before-selection}
	}\\
	\subfloat[After RFECV Selection]{
		\includegraphics[width=.9\linewidth]{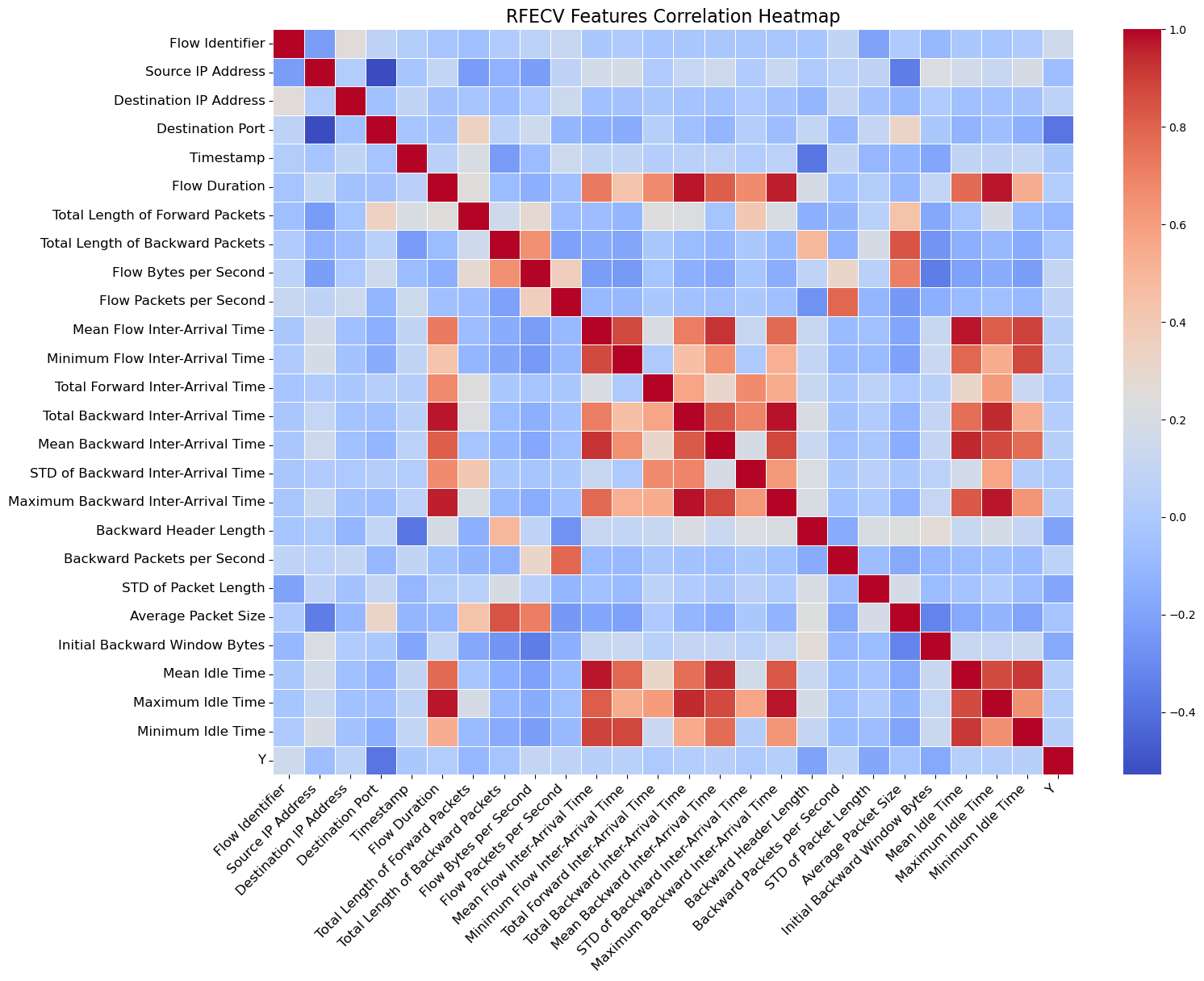}
		\label{fig:after-rfecv}
	}\\
	\subfloat[After Chi2 Selection]{
		\includegraphics[width=.9\linewidth]{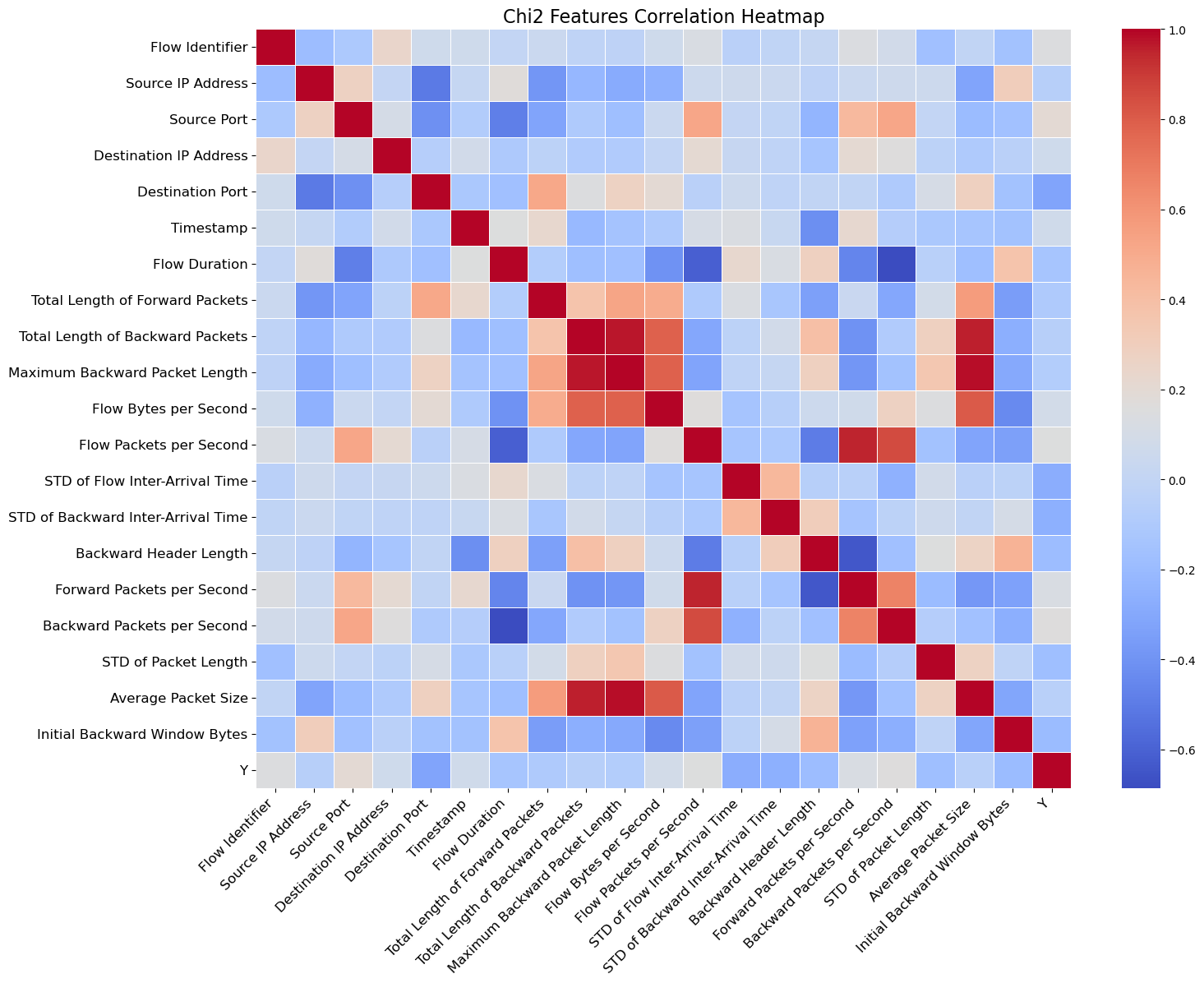}
		\label{fig:after-chi2}
	}
	\caption{Heatmaps illustrating feature correlations before and after applying RFECV and Chi2 feature selection methods.}
	\label{fig:feature-selection-impact}
\end{figure}

\noindent
\textbf{Feature Selection Impact}\\
Feature selection methods show varying impacts across the evaluated models, highlighting the importance of selecting suitable techniques for preprocessing. Initially, the dataset contained 31 features, with several showing redundancy due to strong inter-feature correlations, as depicted in Figure~\ref{fig:feature-selection-impact}-a. 

The RFECV method significantly reduced the number of features, selecting anywhere from 1 to 25 features depending on the normalization and transformation techniques used. For instance, Figure~\ref{fig:feature-selection-impact}-b illustrates a scenario where RFECV retains 25 features, yielding acceptable performance in certain cases. However, combining min-max scaling with the Yeo-Johnson transformation caused some features to become highly redundant, prompting RFECV to retain only a single feature in models 2, 14, and 26. This aggressive feature reduction negatively impacted the autoencoder model since autoencoders encode and decode an input vector; excessive feature reduction can hamper reconstruction accuracy and thus reduce anomaly detection performance, as shown in model~14, causing the accuracy to drop to 51.88\%. The autoencoder's ability to reconstruct input data was compromised, thereby decreasing its anomaly detection performance. Conversely, GBoosting and RNN-LSTM were more robust to the smaller feature sets, maintaining strong performance despite the reduction.

In contrast, the Chi2 method requires specifying the number of features to retain and also mandates non-negative input values, necessitating min-max scaling beforehand. Therefore, Chi2 was configured to select 20 features to balance performance and dimensionality. Figure~\ref{fig:feature-selection-impact}-c shows that Chi2 reduced redundant features while preserving important features. For instance, the GBoosting model (model~25) maintained high accuracy (99.84\%) with this reduced feature set, highlighting Chi2's ability to streamline the feature space without adversely impacting performance.

Overall, while RFECV can inadvertently remove critical features that negatively affect models sensitive to reduced representations, Chi2, on the other hand, provided a more balanced approach, simplifying the feature set without adversely impacting performance.

%=====================================================================================%
\subsection{Performance Analysis by ML Model}
%=====================================================================================%
The analysis below evaluates each machine learning model individually, highlighting key performance trends, sensitivities to preprocessing techniques, and practical implications for IoT anomaly detection.

\noindent
\textbf{GBoosting Model}\\
GBoosting consistently showed superior performance, achieving near-perfect accuracy (99.84\%) and F1-score (99.92\%) across preprocessing configurations. Its robustness and minimal sensitivity to preprocessing choices suggest it's highly reliable and adaptable for anomaly detection in IoT contexts.

\noindent
\textbf{RNN-LSTM Model}\\
RNN-LSTM achieved high accuracy (up to 99.52\%), particularly benefiting from z-score normalization, underscoring its ability to capture temporal dependencies effectively when data is standardized. However, performance was sensitive to feature selection choices, indicating a simpler pipeline without feature selection is preferable. While model 10 provided optimal performance, model 9 (using RFECV) presented a reasonable trade-off, slightly reducing the number of features for improved interpretability.

\noindent
\textbf{Autoencoder Model}\\
Autoencoder models demonstrated excellent recall (up to 99.32\%), making them valuable for scenarios prioritizing minimal false negatives. However, overall accuracy was comparatively lower (94.08\%), and autoencoders proved particularly sensitive to preprocessing and feature selection decisions. Specifically, RFECV significantly reduced accuracy, highlighting the need for cautious application of feature selection methods.

%-------------------------------------------------------------------------------------%
\subsection{Summary of Key Findings}
%-------------------------------------------------------------------------------------%
Across multiple experiments, GBoosting consistently achieved near-perfect accuracy, underscoring its reliability even with minimal preprocessing. RNN-LSTM performance was notably enhanced by z-score normalization but proved sensitive to feature selection. Autoencoders exhibited strong recall but were especially vulnerable to aggressive feature reduction. From these observations, the following general guidelines can assist in designing effective anomaly detection solutions for IoT data:

\begin{itemize}
    \item \textbf{Prioritize Simpler Preprocessing Pipelines:}
    When the dataset does not exhibit extreme skewness, using basic normalization (e.g., z-score) and minimal feature engineering tends to deliver robust results. More complex transformations or feature reductions should be reserved for cases where substantial noise, redundancy, or skewness significantly hinders performance.

    \item \textbf{Align Normalization with Data Characteristics:}
    Datasets featuring outliers or large numeric ranges often benefit most from z-score normalization, as it centers and scales features to handle outliers effectively. If relative feature ratios are crucial and outliers are minimal, min-max scaling remains a viable option.

    \item \textbf{Apply Feature Selection Judiciously:}
    While techniques such as Chi-square and RFECV can reduce dimensionality and improve interpretability, they may also discard essential features, particularly in models like autoencoders that rely on rich input representations. Always verify the net effect on performance metrics before finalizing a feature selection strategy.

    \item \textbf{Evaluate Multiple Models Where Possible:}
    Although a single model (e.g., GBoosting) can excel in many scenarios, specific IoT contexts--such as those requiring strict real-time operations or those with pronounced sequential patterns--may benefit from specialized architectures like RNN-LSTMs. Ultimately, testing multiple models on the target dataset confirms which approach best addresses both performance and deployment needs.
\end{itemize}

Overall, these guidelines emphasize the importance of aligning model choice and preprocessing decisions with the dataset’s inherent properties, deployment constraints, and desired performance trade-offs (e.g., prioritizing recall vs. overall accuracy). By starting with relatively simple pipelines and iteratively refining them in response to observed limitations, practitioners can more effectively tailor anomaly detection strategies to diverse IoT environments.

%%%%%%%%%%%%%%%%%%%%%%%%%%%%%%%%%%%%%%%%%%%%%%%%%%%%%%%%%%%%%%%%%%%%%%%%%%%%%%%%%%%%%%%
%%%%%%%%%%%%%%%%%%%%%%%%%%%%%%%%%%%%%%%%%%%%%%%%%%%%%%%%%%%%%%%%%%%%%%%%%%%%%%%%%%%%%%%  
\section{Conclusion and Future Work}\label{sec:conclusion}
%%%%%%%%%%%%%%%%%%%%%%%%%%%%%%%%%%%%%%%%%%%%%%%%%%%%%%%%%%%%%%%%%%%%%%%%%%%%%%%%%%%%%%%
%%%%%%%%%%%%%%%%%%%%%%%%%%%%%%%%%%%%%%%%%%%%%%%%%%%%%%%%%%%%%%%%%%%%%%%%%%%%%%%%%%%%%%%
This paper introduced a multi-step evaluation framework for IoT anomaly detection that systematically incorporates data preprocessing, transformation, and feature selection. Three primary machine learning models—RNN-LSTM, autoencoder, and Gradient Boosting—were evaluated using the IoTID20 dataset. The results showed that Gradient Boosting consistently achieved high accuracy under various preprocessing configurations, indicating robust performance in diverse feature distributions. RNN-LSTM provided comparable results while effectively capturing temporal dependencies, making it highly suited for time-series-based IoT data. Autoencoders displayed slightly lower overall accuracy but excelled in recall, suggesting their potential in scenarios where minimizing false negatives is critical.

\noindent
\textbf{Future Work:}\\
To further enhance and validate the proposed framework, future research can focus on:
\begin{itemize}
    \item \textbf{Additional Datasets and Scenarios:} Evaluating the framework on different IoT datasets, including those with novel attack profiles or resource-constrained environments, to test its generalization.
    \item \textbf{Extended Preprocessing Techniques:} Incorporating more transformation methods (e.g., Box-Cox), dimensionality reduction algorithms (e.g., PCA), and feature engineering strategies to examine their impact on anomaly detection performance.
    \item \textbf{Advanced Model Architectures and Hyperparameter Tuning:} Exploring a broader range of neural network architectures (e.g., deeper LSTM layers, CNN-LSTM hybrids) and leveraging Bayesian optimization or other automated approaches for hyperparameter tuning. This could further improve accuracy and reduce manual trial-and-error.
    \item \textbf{Diverse ML Methods:} Investigating additional machine learning algorithms—such as random forests, one-class SVMs, or ensemble methods--under the same pipeline to identify optimal solutions for specific IoT use cases.
    \item \textbf{Computational Efficiency and Real-Time Constraints:} Assessing the run-time performance and resource consumption of the proposed models, especially those using deep learning architectures like RNN-LSTM, to ensure practical feasibility for real-time anomaly detection in resource-limited IoT deployments.
    \item \textbf{Handling Class Imbalance:} Exploring various strategies—such as oversampling minority classes (e.g., SMOTE), undersampling majority classes, cost-sensitive learning, and threshold optimization—to address class imbalance. Evaluating these techniques would enhance anomaly detection performance, particularly for minority class instances typically found in IoT datasets.
\end{itemize}

Overall, the findings highlight the importance of selecting an appropriate combination of preprocessing steps and learning algorithms for high-fidelity anomaly detection in IoT networks. The proposed framework offers a solid foundation for ongoing research and real-world applications, where continuous experimentation and refinement can further enhance security in increasingly interconnected environments.

%%%%%%%%%%%%%%%%%%%%%%%%%%%%%%%%%%%%%%%%%%%%%%%%%%%%%%%%%%%%%%%%%%%%%%%%%%%%%%%%%%%%%%%
%%%%%%%%%%%%%%%%%%%%%%%%%%%%%%%%%%%%%%%%%%%%%%%%%%%%%%%%%%%%%%%%%%%%%%%%%%%%%%%%%%%%%%%

%%%%%%%%%%%%%%%%%%%%%%%%%%%%%%%%%%%%%%%%%%%%%%%%%%%%%%%%%%%%%%%%%%%%%%%%%%%%%%%%%%%%%%%
%%%%%%%%%%%%%%%%%%%%%%%%%%%%%%%%%%%%%%%%%%%%%%%%%%%%%%%%%%%%%%%%%%%%%%%%%%%%%%%%%%%%%%%  
% \section*{Acknowledgment}
%%%%%%%%%%%%%%%%%%%%%%%%%%%%%%%%%%%%%%%%%%%%%%%%%%%%%%%%%%%%%%%%%%%%%%%%%%%%%%%%%%%%%%%
%%%%%%%%%%%%%%%%%%%%%%%%%%%%%%%%%%%%%%%%%%%%%%%%%%%%%%%%%%%%%%%%%%%%%%%%%%%%%%%%%%%%%%%
% This research has been supported by NSERC under grant RGPIN-2018-06222.

%%%%%%%%%%%%%%%%%%%%%%%%%%%%%%%%%%%%%%%%%%%%%%%%%%%%%%%%%%%%%%%%%%%%%%%%%%%%%%%%%%%%%%%
%%%%%%%%%%%%%%%%%%%%%%%%%%%%%%%%%%%%% References %%%%%%%%%%%%%%%%%%%%%%%%%%%%%%%%%%%%%%
%%%%%%%%%%%%%%%%%%%%%%%%%%%%%%%%%%%%%%%%%%%%%%%%%%%%%%%%%%%%%%%%%%%%%%%%%%%%%%%%%%%%%%% 
\bibliographystyle{IEEEtran}
\bibliography{07_references.bib}

%%%%%%%%%%%%%%%%%%%%%%%%%%%%%%%%%%%%%%%%%%%%%%%%%%%%%%%%%%%%%%%%%%%%%%%%%%%%%%%%%%%%%%%
%%%%%%%%%%%%%%%%%%%%%%%%%%%%%%%%%%%%%%%%%%%%%%%%%%%%%%%%%%%%%%%%%%%%%%%%%%%%%%%%%%%%%%%

\end{document}